\documentclass[10pt,twocolumn,letterpaper]{article}

\usepackage{cvpr}
\usepackage{times}
\usepackage{epsfig}
\usepackage{graphicx}
\usepackage{amsmath}
\usepackage{amssymb}
\usepackage{multirow}
\usepackage{booktabs}
\usepackage{epsfig}
\usepackage[table,xcdraw]{xcolor}

\newcommand{\eat}[1]{}

\usepackage[pagebackref=true,breaklinks=true,letterpaper=true,colorlinks,bookmarks=false]{hyperref}

 \cvprfinalcopy 

\ifcvprfinal\fi

\renewcommand{\paragraph}{\textbf}
\newcommand{\zx}{\begin{equation}}
\newcommand{\xz}{\end{equation}}
\expandafter\def\expandafter\normalsize\expandafter{%
	\normalsize\setlength\abovedisplayskip{3pt}}
\expandafter\def\expandafter\normalsize\expandafter{%
	\normalsize\setlength\belowdisplayskip{3pt}}

\begin{document}

\title{Deep Region Hashing for Efficient Large-scale Instance Search from Images}

\author{Jingkuan Song\\
Columbia University\\
{\tt\small jingkuan.song@gmail.com}
\and
Tao He\\
University of Electronic Science\\ and Technology of China\\
{\tt\small tao.he@uestc.edu.cn}
\and
Lianli Gao\\
University of Electronic Science\\ and Technology of China\\
{\tt\small lianli.gao@uestc.edu.cn}
\and
Xing Xu\\
University of Electronic Science\\ and Technology of China\\
{\tt\small xing.xu@uestc.edu.cn}
\and
Heng Tao Shen\\
University of Electronic Science\\ and Technology of China\\
{\tt\small shenhengtao@hotmail.com}
}
\maketitle

\begin{abstract}
Instance Search (INS) is a fundamental problem for many applications, while it is more challenging comparing to traditional image search since the relevancy is defined at the instance level.
 Existing works have demonstrated the success of many complex ensemble systems that are typically conducted by firstly generating object proposals, and then extracting handcrafted and/or CNN features of each proposal for matching. 
However, object bounding box proposals and feature extraction are often conducted in two separated steps, thus the effectiveness of these methods collapses. Also, due to the large amount of generated proposals, matching speed becomes the bottleneck that limits its application to large-scale datasets.
To tackle these issues, in this paper we propose an effective and efficient Deep Region Hashing (DRH) approach for large-scale INS using an image patch as the query. Specifically, DRH is an end-to-end deep neural network which consists of object proposal, feature extraction, and hash code generation. 
DRH shares full-image convolutional feature map with the region proposal network, thus enabling nearly cost-free region proposals. Also, each high-dimensional, real-valued region features are mapped onto a low-dimensional, compact binary codes for the efficient object region level matching on large-scale dataset.
Experimental results on four datasets show that our DRH can achieve even better performance than the state-of-the-arts in terms of MAP, while the efficiency is improved by nearly 100 times.
\end{abstract}

\section{Introduction}
\eat{\textcolor{red}{why choose cnn features}}
Large-scale instance search (INS) is to efficiently retrieve the images containing a specific instance in a large scale image dataset, giving a query image of that instance. It has long been a hot research topic due to the many applications such as image classification and detection \cite{vedaldi2012sparse,DBLP:journals/corr/WangZSSS16}.

\begin{figure}[t]
\centering
\includegraphics[width=3.5in]{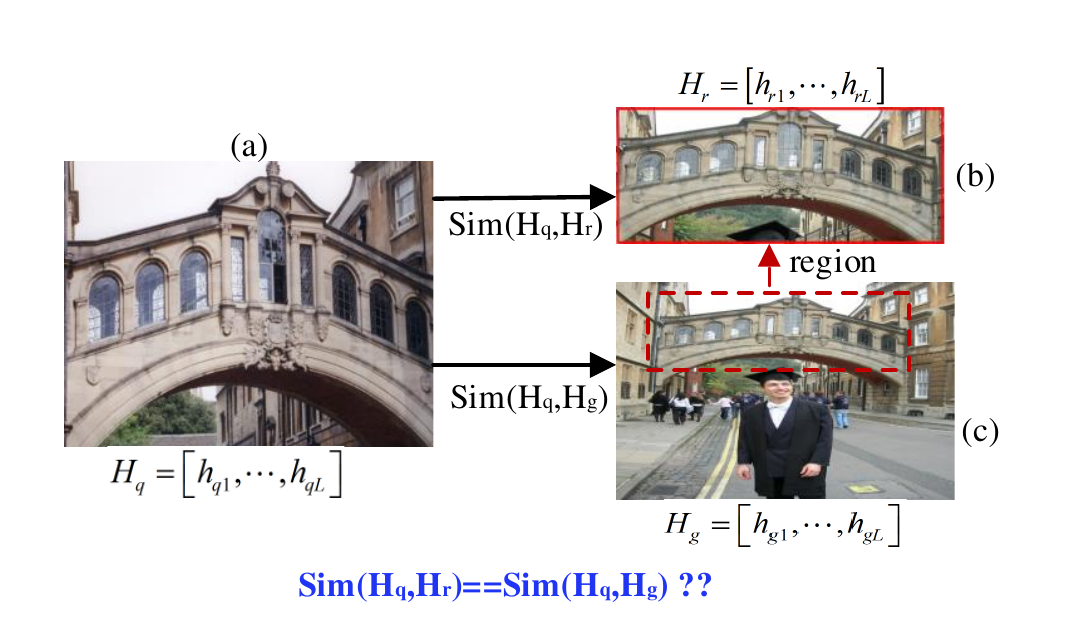}
\caption{Purely using global hash codes to calculate the similarity between query instance and database images is not accurate enough. (a) is a query instance with a deep hash code representation $H_q$; (c) is a database image with a deep global region hash code $H_g$ by taking the whole image as an region; (b) is an region pooled from (c) and with a deep region hash code $H_l$. However, it is clear that conduct INS search only using global region (c) information is inaccurate and the deep region hashing should be considered as well. Thus, in this paper, we aim to learn a deep region hashing to support fast and efficient INS.}
\label{fig:intro}
\end{figure}

Early research \cite{Philbin-cvpr07} usually extracts image-level handcrafted features such as color, shape and texture to search a user query. Recently, due to the development of deep neural networks, using the outputs of last fully-connected network layers as global image descriptors to support image classification and retrieval have demonstrated their advantage over prior state-of-the-art~\cite{DBLP:journals/corr/NgYD15}. More recently, we have witnessed the research attention shifted from features extracted from the fully-connected layers to deep convolutional layers \cite{DBLP:journals/corr/BabenkoL15}, and it turns out that convolutional layer activations have superior performance in image retrieval. 
However, using image-level features may fail to search instances in images. As illustrated in Fig. \ref{fig:intro}, sometimes a target region is only a small portion of a database image. Directly comparing the query image and a whole image in the database may result an inaccurate matching. Therefore, it is necessary to consider the local region information.

\eat{\textcolor{red}{why need region}}
Recently, many successful INS approaches \cite{cvpr20146909666,DBLP:journals/corr/MohedanoSMMON16} were proposed by combining fast filtering with more computationally expensive spatial verification and query expansion.
More specifically, all images in a database are firstly ranked according to their distances to the query, and then a spatial verification is applied to re-rank the search results, and finally a query expansion is followed to improve the search performance.
An intuitive way for spatial verification is the sliding window strategy. It generates sliding windows at different scales and aspect ratios over an image. Each window is then compared to the query instance in order to find the optimal location that contains the query.
A more advanced strategy is to use object proposal algorithms \cite{UijlingsIJCV2013,girshick14CVPR}. It evaluates many image locations and determines whether they contain the object or not. Therefore, the query instance only need to compare with the object proposal instead of all the image locations.
Existing works \cite{cvpr20146909666,7298613} have demonstrated the success of many complex ensemble systems that are conducted by firstly generating object bounding box proposals and then extracting handcrafted and/or CNN features for each bonding box. Because the region detection and feature extraction are conducted in two separate steps, the detected regions may not be the optimal one for feature extraction, leading to suboptimal results. 

Faster R-CNN \cite{renNIPS15fasterrcnn} introduces a region proposal network which integrates feature extraction and region proposal in a single network. The detection network shares the full-image convolutional features, thus enabling nearly cost-free region proposals. Inspired by this, if a neural network can simultaneous extract the convolutional features and propose some object candidates, we can perform instance search on these candidates. However, efficiency is a big issue for Faster R-CNN. 
The successfully selective search and RPN are still generating hundreds to thousands of candidate regions. If a dataset has $N$ images and each image contains $M$ regions, each region is represented by a $d$-dimensional feature vector, then an exhaustive search for each query requires $N \times M \times d$ operation to measure the distances to the candidate regions.
Therefore, despite taking the advantages of deep convlutional layer activations and region proposals, the expensive computational cost for nearest neighbors search in high-dimensional data limits their application to large scale datasets.
\eat{\textcolor{red}{The challenges and why hashing} }
\eat{
The goal of our paper is required to be performed fast so that users can interactively browse the dataset.  the big challenge for INS here is the search efficiency and memory cost. 
The successfully selective search and RPN are still evaluating thousands of candidate regions. If a dataset has $N$ images and each image contains $M$ regions, each region is represented by a $d$-dimensional feature vector, then an exhaustive search for each query requires $N \times M \times d$ operation to measure the distance between vectors. 
A common approach for accelerating INS search is indexing, however as the dimensional grows, the probability of similar images are assigned to different clusters increases, thus the efficiency of these methods collapses. Thus, the big challenge here is to improve the INS efficiency and decrease memory requirements without sacrificing accuracy. Therefore, we need to solve the problem of region similarity search since it calculating distance between the query vector and database vectors is costly and often computationally infeasible. Hashing is a powerful technique for large scale visual search and the basic idea of hashing-based approach is to transform a high-dimensional image descriptor to a low-dimensional representation, or equivalently a short code consisting of a sequence of bits.
}
\eat{\textcolor{red}{why our methods}}

In this paper we propose a fast and effective Deep Region Hashing (DRH) approach for visual instance search in large scale datasets with an image patch as the query.
Specifically, DRH is an end-to-end deep neural network which consists of layers for feature extraction, object proposal, and hash code generation. 
DRH can generate nearly cost-free region proposals because the region proposal network shares full-image convolutional feature map. Also, the hash code generation layer learns binary codes to represent large scale images and regions to accelerate INS.
It is worthwhile to highlight the following aspects of DRH:
\begin{itemize}
	\item We propose an effective and efficient end-to-end Deep Region Hashing (DRH) approach, which consists of object proposal, feature extraction, and hash code generation, for large-scale INS with an image patch as the query. Given an image, DRH can automatically generate the hash codes for the whole image and the object candidate regions.
	\item We design different strategies for object proposals based on the convolutional feature map. The region proposal is nearly cost-free, and we also integrate hashing strategy into our approach to enable efficient INS on large-scale datasets.
	\item Extensive experimental results on four datasets show that our generated hash codes can achieve even better performance than the state-of-the-arts using real-valued features in terms of MAP, while the efficiency is improved by nearly 100 times.
\end{itemize}


\begin{figure*}[t]
\centering
\includegraphics[width=0.95\linewidth,height=0.85\columnwidth]{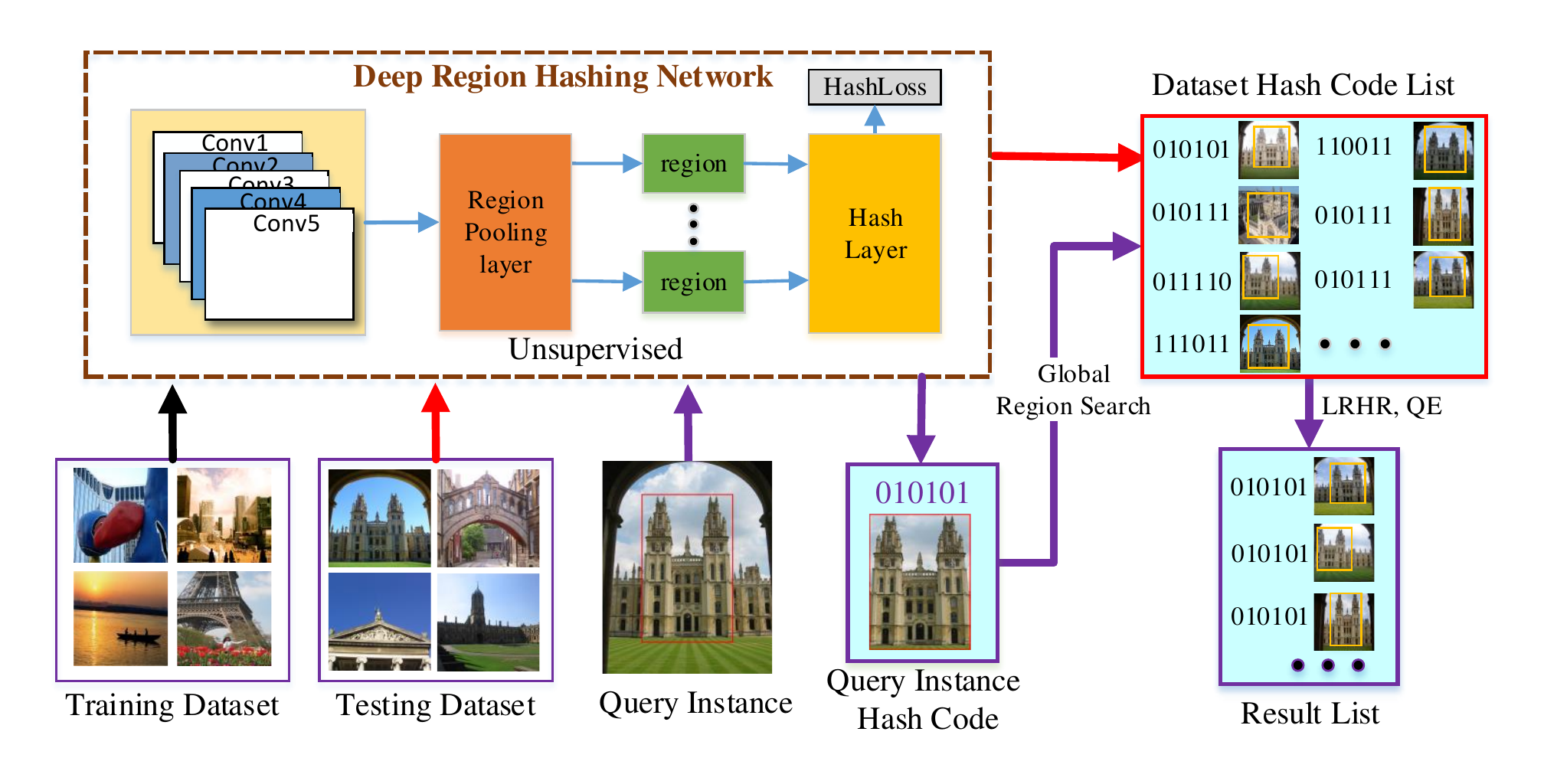}
\caption{The overview of DRH. It contains two parts, i.e., deep region hashing network part and instance search part. The first part trains a network for feature extraction, object proposal and hash code generation, while the second part utilizes DRH to get the query results.}
\label{fig:framework}
\end{figure*}

\section{Related Work}
Our work is closed related to instance retrieval, object proposal and hashing. We brief review the related work for each of them.

\textbf{Instance retrieval.} Most approaches in INS rely on the bag-of-words (BoW) model \cite{6619055,Philbin-cvpr07,DBLP:journals/corr/MohedanoSMMON16}, which is one of the most effective content-based approaches for large scale image and visual object retrieval. Typically, they are based on local features such as RootSIFT \cite{cvpr20146909666} and few are based on CNN features. For instance, \cite{DBLP:journals/corr/MohedanoSMMON16} proposes to conduct INS by encoding the convolutional features of CNN using BoW aggregation scheme. More recently, research has shown that VLAD \cite{5540039} and fisher vector \cite{5540009} performs better than BoW \cite{cvpr20146909666,DBLP:journals/corr/KalantidisMO15,7298613}. Specifically, BoW encodes a local representation by indexing the nearest visual word, while VLAD and fisher vector quantize the local descriptor with an extra residual vector obtained by subtracting the mean of the visual word or a Gaussian component fit to all observations \cite{cvpr20146909666,7298613}. However, embedding methods have drawbacks, e.g., inhibiting true positive matches between local features, consisting of lots of parameters and suffering from overfitting \cite{7298613}.

Several works are proposed to go beyond handcrafted features for image retrieval and INS. \cite{Krizhevsky_imagenetclassification,DBLP:journals/corr/ConneauSBL16} trained deep CNNs and adopted features extracted from the fully connected layers to enhance image retrieval. Recently, research attention shifted from extracting deep fully-connected features to deep convolutional features \cite{DBLP:journals/corr/RazavianSMC14,wang2015visual,Cimpoi15}. For instance, \cite{DBLP:journals/corr/BabenkoL15} aggregates local deep features to produce compact global descriptors for image retrieval. \cite{DBLP:journals/corr/RazavianSMC14} provides an extensive study on the availability of image representations based on convolutional networks for INS. In addition, \cite{DBLP:journals/corr/BabenkoSCL14} demonstrates that the activations invoked by an image within the top layers of a large convolutional neural network provide a high-level descriptor of the visual content of the image. Also, PCA-compressed neural codes outperform compact descriptors compressed on handcrafted features such as SIFT.  In addition, \cite{Salvador_2016_CVPR_Workshops} explores the suitability for INS of image and region-wise representations pooled from the trained Faster R-CNN or fine-turned Faster R-CNN with supervised information by combining re-ranking and query expansions. 

 
\textbf{Object proposals.} Object proposals have received much attention recently. It can be broadly categorized into two categories: super-pixels based methods (e.g., Selective Search \cite{UijlingsIJCV2013} and  Multiscale Combinatorial Grouping \cite{PABMM2015}) and sliding window based approaches (e.g., EdgeBoxes and Faster R-CNN \cite{renNIPS15fasterrcnn}). More specifically, selective search adopts the image structure for guiding sampling process and captures all possible locations by greedily merging superpixels based on engineered low-level features. Due to its good performance, it has been widely used as an initial step for detection such as R-CNN \cite{girshick14CVPR} and INS task \cite{cvpr20146909666}. Even though MCG achieves the best accuracy, it takes about 50 seconds for each image to generate region proposals. By contrast, Faster R-CNN proposed a region propsoal network, which takes an image as input, shares a common set of convolutional layers with object detection network, and finally outputs a set of rectangular objects proposals, each with an abjectness scores. This region proposal network is trained on the PASCAL VOC dataset and achieves high accuracy. A comprehensive survey and evaluation protocol for object proposal methods can be found in~\cite{Chavali_2016_CVPR}.

\textbf{Hashing.} Existing hashing methods can be classified into two categories: unsupervised hashing \cite{wang2012semi,Liong_2015_CVPR} and supervised hashing \cite{wang2012semi,6619050,Song_2015_ICCV,Liong_2015_CVPR,Zhao_2015_CVPR}. A comprehensive survey and comparison of hashing methods are represented in \cite{DBLP:journals/corr/WangZSSS16}. CNN have achieved great success in various tasks such as image classification, retrieval, tracking and object detection \cite{Chavali_2016_CVPR,Cimpoi15,DBLP:journals/corr/BabenkoSCL14,DBLP:journals/corr/ConneauSBL16,DBLP:journals/corr/NgYD15,wang2015visual}, and there are few hashing methods that adopts deep models. For instance, Zhao \textit{et. al.}\cite{Zhao_2015_CVPR} proposed a deep semantic ranking based method for learning hash functions that preserve multilevel semantic similarity between multi-label images. Lin \textit{et. al.} \cite{Lin_2015_CVPR} propose an effective semantics-preserving Hashing method, which transforms semantic affinities of training data into a probability distribution.  To date, most of the deep hashing methods are supervised and they tried to leverage supervised information (e.g., class labels) to learn compact bitwise representations, while few methods are focused on unsupervised methods, which used unlabeled data to learn a set of hash functions. However, the proposed deep unsupervised hashing method are focused on the global image presentations without considering the locality information which we believe to be important for INS.


\section{Methodology}
This paper explores instance search from images using hash codes of image regions detected by an object detection CNN or sliding window. In our setup, query instances are defined by a bounding box over the query images.
The framework of our DRH-based instance search is shown in Fig.\ref{fig:framework}, and it has two phases, i.e., offline DRH training phase and online instance search phase.
In this section, we describe these two phases in details.

 \subsection{Deep Region Hashing}
As is shown in Fig.~\ref{fig:framework}, the DRH training phase consists of three components, i.e., CNN-based feature map generation, region proposals and region hash code generation. Next, we illustrate each of them.

\subsubsection{CNN-based Representations}
As mentioned above, we focus on leveraging convolutional networks for hash learning. We adopt the well known architecture in \cite{DBLP:journals/corr/ConneauSBL16} as our basic framework. As shown in {Fig.~\ref{fig:framework}}, our network has $5$ groups of convolution layers, $4$ max convolution-pooling layers, $1$ Region of Interest Pooling layer (RoI) and $1$ hashing layer. Following \cite{DBLP:journals/corr/ConneauSBL16}, we use $64$, $128$, $256$, $512$, $512$ filters in the $5$ groups of convolutional layers, respectively. More specifically, our network firstly processes an arbitrary image with several convolutional (conv) and max pooling layers to produce a conv feature map. Next, for each region proposal, a region of interest (RoI) pooling layer extracts a fixed-length (i.e., 512) feature from the feature map. Each feature map is then fed into the region hash layer that finally outputs hash codes representing the corresponding RoI.

\subsubsection{Region Pooling Layer}
There are two ways to generate region proposals: 1) sliding window (the number of region proposal is determined by a sliding window overlaping parameter $\lambda$ ) and 2) the Region Proposal Network (RPN) proposed by Fater RCNN \cite{renNIPS15fasterrcnn}. It provides a set of object proposals, and we use them as our region proposals by assuming that the query image always contains some objects. 
\begin{itemize}
\item \textbf{Sliding window.} Given a raw image with the size of $W \times H$, our network firstly generates a conv feature map ${W_c} \times {H_c}$, where $W$, ${W_c}$ and $H$, ${H_c}$ are the width and height of the original image and conv map, respectively. Next, we choose windows of any possible combinations of width $\left\{ {{W_c},\frac{{{W_c}}}{2},\frac{{{W_c}}}{3}} \right\}$ and height $\left\{ {{H_c},\frac{{{H_c}}}{2},\frac{{{H_c}}}{3}} \right\}$. We use a sliding window strategy directly on the conv map with overlap in both directions. The percentage of overlap is measured by a overlapping parameter $\lambda$. {Next, we utilize a simple filtering strategy to discard those windows. If $\frac{W}{H} \le th$, we discard the scale ${\frac{{{W_c}}}{3}}$.  When $\frac{H}{W} \le th$, we discard ${\frac{{{H_c}}}{3}}$, otherwise we keep the original settings.}
When the width and height are set to $W_c$ and $H_c$, we can generate an global region image descriptor, which will be used as the input for global region hash code generation.

\item \textbf{RPN network.} Faster R-CNN \cite{renNIPS15fasterrcnn} proposed a Region Proposal Network (RPN), which takes conv feature map as input and outputs a set of rectangular object proposals, each with an abjectness scores. Given a conv feature map ${W_c} \times {H_c}$, it generates ${W_c} \times {H_c}  \times 9$ (typically $216,00$ regions). Specifically, the RPN proposed by Faster R-CNN is able to propose object regions and extracts the convolutional activations for each of them. Therefore, for each region proposal, it can compose a descriptor by aggregating the activations of that window in the RoI pooling layer, giving raise to the region-wise descriptors. In order to conduct image-level operation, we obtain a global region by setting the whole conv map as a region. To conduct instance-level search, we obtain local regions which have smaller size than the global region.
\end{itemize}

After region proposal generation either adopting sliding window based approach or RPN approach, for each image we obtain a set of regions which include a global region and several local regions. Next, we apply max pooling to the corresponding region window on the conv map to generate ROI feature for each region.

\subsubsection{Region Hashing Layer}
For instance search, directly using RoI features to compare the query feature with all the ROI features is effective and it should work well in small-scale dataset. However, in terms of large-scale dataset, this approach is computationally expensive and requires large memory since it involves huge vector operations. In this paper, we aim to improve INS search accuracy and efficiency. Inspired by the success of hashing in terms of search accuracy, search time cost and space cost, we propose a region hashing layer to generate hash codes for the region proposals.

Here, we convert each RoI feature to  a set of binary hash code. Given a RoI feature ${x_r} \in {\mathbb{R}^{1 \times 512}}$, the latent region hash layer output is ${h_r} = \sigma ({W_r}{x_r} + b_r)$, where $W_r$ is the hash function and $b_r$ is the bias. Next, we perform the binary hash for the output ${h_r} \in {\mathbb{R}^{{1 \times L_r}}}$ to obtain a binary code ${y_r} = sgn(\sigma (W_r{x_r} + b_r)) \in \{0,1\}^{L_r}$. To obtain a qualified binary hash code, we propose the following optimization problem:
\begin{eqnarray}
\mathop {\min {\rm{ }}}\limits_{W_r,b_r} \ell =  \frac{1}{2}g\left( {{y_r},{h_r}} \right) -  \frac{\alpha}{2} t({h_r}) + \frac{\beta}{2}  r\left( W_r \right) + \frac{\eta}{2}  o\left( b_r \right)
\label{eq.ogl}
\end{eqnarray}
where $g\left( {{y_r},{h_r}} \right)$ is the penalty function to minimize the loss between the RoI feature $h_r$ and $y_r$. $t({h_r}), ~r\left( W_r \right)$ and $o\left( b_r \right) $ are the regularization term for $h_r,~W_r$ and $b_r$. $\alpha,~\beta$ and $\eta$ are parameters. We define $g\left( {{y_r},{h_r}} \right)$ and $t({h_r})$ as bellow:
\begin{eqnarray}
\begin{array}{l}
g\left( {{y_r},{h_r}} \right) = \left\| {{y_r} - {h_r}} \right\|_F^2\\
t({h_r}) = tr\left( {{h_r}{h_r}^T} \right)
\end{array}
\label{eq.obj.hash}
\end{eqnarray}

The regularize terms $r\left( W_r \right)$ and $o\left( b_r \right)$ are defined as:
\begin{eqnarray}
\begin{array}{l}
r\left( W_r \right) = \left\| {{W_r}} \right\|_F^2\\
o\left( {{b_r}} \right) = \left\| {{b_r}} \right\|_F^2
\end{array}
\end{eqnarray}

To solve this optimization problem, we employ the stochastic gradient descent method to learn parameters $W_r,b_r$. The gradient of the objective function in (\ref{eq.obj.hash}) with respect to different parameters are computed as following: 
\begin{align}
\begin{array}{l}
\frac{{\partial \ell  }}{{\partial {W_{\rm{r}}}}} = \left( {({h_r} - {{\rm{y}}_r} - \alpha {h_r})  \odot \sigma '( {W_r}{h_r} + {b_r})} \right){x_r}^T + \beta {W_r}\\
\frac{{\partial \ell }}{{\partial {b_r}}} = ({h_r} - {{\rm{y}}_r} - \alpha {h_r}) \odot \sigma '({W_r}{h_r} + {b_r}) + \eta {b_r}
\end{array}
\label{eq.sol}
\end{align}

where $ \odot $ is element-wise multiplication. The parameters ${{W_{\rm{r}}}}$ and ${b_r}$ are updated with gradient descent algorithm until convergence.

\subsubsection{Network Training}
 We train our models using stochastic gradient descent with momentum, with back-propagation used to compute the gradient of the objective function $\nabla Loss\left( {{W_r},{b_r}} \right)$ with respect to all parameters $W_r$ and $b_r$. More specifically, we randomly initialize the RoI pooling layer and region hashing layer by drawing weights from a zero-mean Gaussian distribution with standard deviation $0.01$. The shared convolutional layers are initialized by pre-training a model \cite{DBLP:journals/corr/RadenovicTC16}. Next, we tune all the layers of the RoI layer and the hashing layer to conserve memory. In particular, for RPN training we fix the convolutional layers and only update the RPN layers parameters. The pascal\_voc 2007 dataset is used to train the RPN and for this study we choose the Top \~300 regions as input to the region hash layer. For sliding window based region proposal, we use different scales to directly extract regions and then input them into the region hash layer. At this stage, the parameters for conv layers and region pooling layers are obtained. Next, we fix the parameters of conv layers and the region pooling layer to train the region hash layer. In addition, learning rate is set to $0.01$ in the experiments. The parameters $\alpha$, $\beta $, $\eta$ were empirically set as $100$, $0.001$ and $0.001$ respectively. 
 
\subsection{Instance Search using DRH}
This section describes the INS pipeline by utilizing DRH approach. This pipeline consists of a Global Region Hashing (gDRH) search, Local Region Hashing (lDRH) re-ranking and two Region Hashing Query Expansions (QE). 

\textbf{Global Region Hashing Search (gDRH).} It is conducted by computing the similarity between the hash code of a query and the global region hash codes of dataset images. For each item in the dataset, we use the hash codes generated by the whole image region. Next, the image query list is sorted based on the hamming distance of dataset items to the query. After gDRH, we choose top $M$ images to form the ${1^{st}}$ ranking list ${X^{rank}} = \left\{ {x_1^{rank}, \cdots ,x_M^{rank}} \right\}$.

\textbf{Local Region Hashing Re-search (lDRH).} After we obtained the ${X^{rank}}$, we perform the local region hashing search by computing the hamming distance between query instance hash code and hash codes of each local region in the top $M$ images. To clarify, the hashing query expansions introduced below is optional. For each image, the distance score is calculated by using the minimal hamming distances between the hash code of query instance and all the local region hash codes within that image. Finally, the images are re-ordered based on the distances and we get the final query list ${X_1^{rank'}} = \left\{ {x_1^{rank'}, \cdots ,x_M^{rank'}} \right\}$.

\textbf{Region Hashing Query Expansions (QE).} In this study, we further investigate two query expansion strategies based on the global and local hash codes.
\begin{itemize}
\item \textit{Global Region Hashing Query expansion (gQE).} After the Global Region Search, we conduct the global hashing query expansion, which is applied after the gDRH and before the lDRH. Specifically, we firstly choose the global hash code of the top $q$ images within the ${X^{rank}}$ to calculate similarities between each of them with each global hash code of the ${X^{rank}}$ images. For $x_i^{rank'}$, the similarity score is computed by using the max similarity between $x_i^{rank'}$ and the top $q$ images' global region hash codes. Next, the ${X^{rank}}$ is re-ordered to ${X^{rankq}} = \left\{ {x_1^{rankq}, \cdots ,x_M^{rankq}} \right\}$ and finally we conduct lDRH on the ${X^{rankq}}$ list.  
 
\item \textit{Local Region Hashing Query Expansion (lQE).} Here, to conduct local region hashing query expansion, initially we need to conduct gDRH and lDRH to get the ${X^{rank'}}$. Here gQE is optional. Next, we get the top $q$ best matched region hash code from images within the ${X^{rank'}}$ list and then compare the $q$ best region hash codes with all the regions of the ${X^{rank'}}$. Thus, we choose the max value as the similarity score for each image and finally the ${X^{rank'}}$ is re-ranked based on those similarity scores.
 \end{itemize}

\section{Experiment}
We evaluate our algorithm on the task of instance search. Firstly, we study the influence of different components of our framework. Then, we compare DRH with state-of-the-art algorithms in terms of efficiency and effectiveness on two standard datasets.
\subsection{Datasets and Evaluation Metric}
We consider two publicly available datasets that have been widely used in previous work. 1) \textbf{Oxford Buildings} \cite{Philbin-cvpr07}. This dataset contains two sub-datasets: \textit{Oxford 5k}, which contains 5,062 high revolutionary ($1024\times 768$) images, including 11 different landmarks (i.e., a particular part of a building), and each represented by several possible queries; and \textit{Oxford 105K}, which combines Oxford 5K with $100,000$ distractors to allow for evaluation of scalability. 2) \textbf{Paris Buildings}~\cite{Philbin08}, which contains two sub-datasets as well: \textit{Paris 6k} including 6,300 high revolutionary ($1024\times 768$) images of Paris landmarks; and \textit{Paris 106 K}, which is generated by adding $100,000$ Flickr distractor images to \textit{Paris 6k} dataset. Compared with Oxford Buildings, it has images representing buildings with some similarities in architectural styles. 
The hashing layer is trained on a dataset which is composed of \textit{Oxford 5K} training examples and \textit{Paris 6k} training examples. 

For each dataset, we follow the standard procedures to evaluate the retrieval performance. In this paper, we use mean Average Precision (mAP) as the evaluation metric.


\begin{figure*}[ht]
\centering
\includegraphics[width=\textwidth,height=1.1\columnwidth]{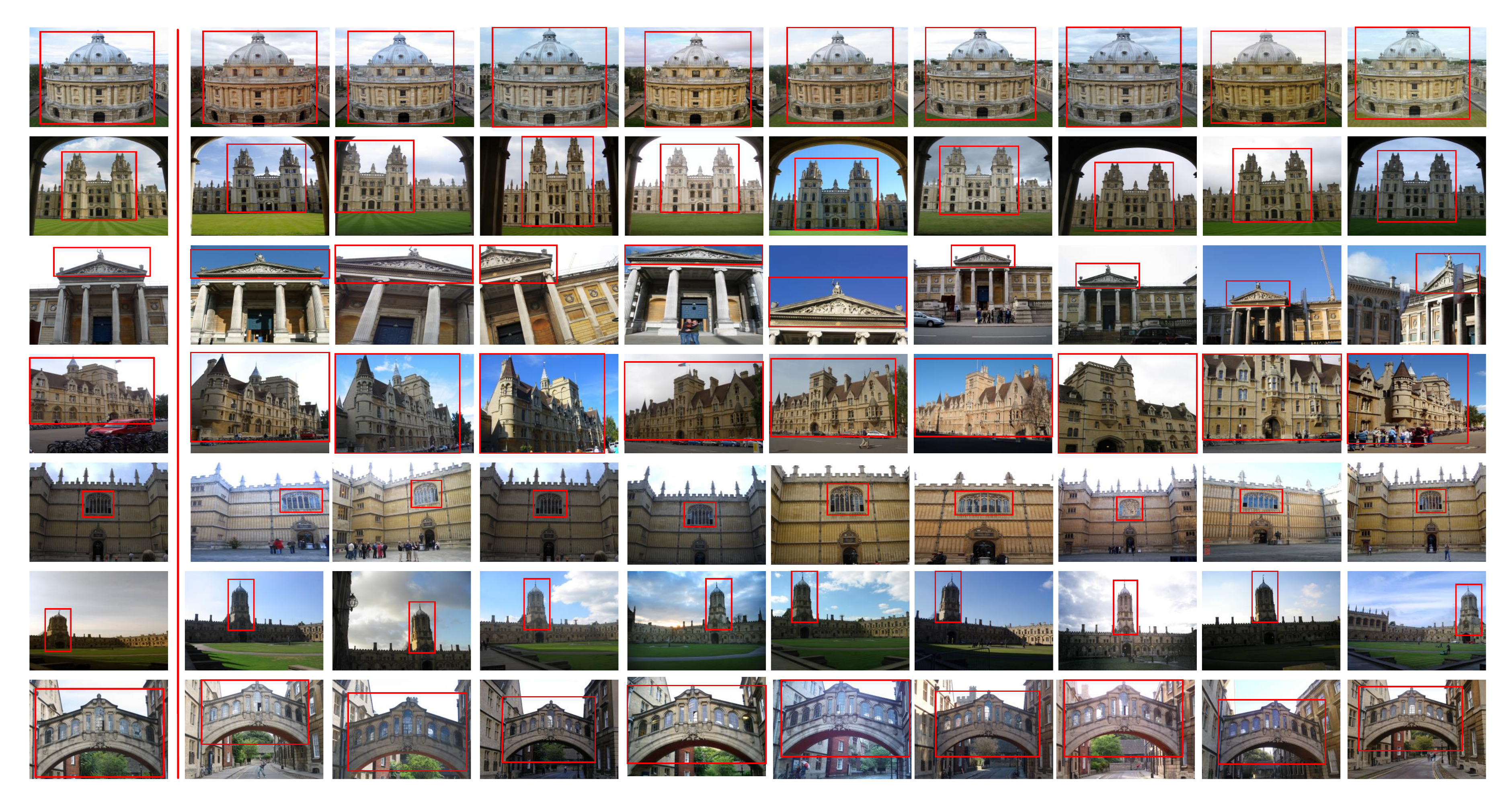}
\caption{The qualitative results of DRH searching on $5$K dataset. $7$ queries (i.e., \textit{redcliffe\_camera, all\_souls, ashmolean, balliol, bodleian, christ\_church and hertford}) and the corresponding search results are shown in each row. Specifically, the query is shown in the left, with selected top $9$ ranked retrieved images shown in the right.}
\label{fig:example}
\end{figure*}

\subsection{Performance using Different Settings}
There are several different settings affecting the performance of our algorithm. To comprehensively study the performance of DRH with different settings, we further study different components of DRH including: 1) Hash code length $L$; 2) region proposal component and its parameters; and  3) query expansion and its settings.

\subsubsection{The Effect of Hash Code Length $L$} 
Firstly, we investigate the influence of hash code length by using gDRH to conduct INS. We report the mAPs of gDRH with different code length $L$, and also the mAP of INS using the max pooling on conv5 feature. The results are shown in Tab.~\ref{hash_code_length}. These results show that with the increase of hash code length $L$ from $128$ to $4096$, the performance of gDRH increases from $42.5\%$ to $78.3\%$ on the \textit{oxford 5k} dataset, while the mAP increases from $53.1\%$ to $81.5\%$ on the \textit{Paris 6k} dataset. Comparing $L=1024$ to $L=4096$, the code length increased to 4 times, but the performance is only improved by $3.5\%$ and $4.2\%$. Therefore, to balance the search efficiency and effectiveness, we use $L=1024$ as the default setting. On the other hand, the best performance of gDRH (i.e., $L=4096$) is slightly lower than that of using max pooling on conv5 features, with $1.4\%$ and $0.9\%$ decreases, respectively. This is due to the information loss of hash codes.

\begin{table}[]
\centering
\caption{The performance (mAP) variance with different code lengths L using gDRH.}
\label{hash_code_length}
\begin{tabular}{|c|c|c|}
\hline
\textbf{Method} & \textbf{oxford 5K} & \textbf{Paris 6K} \\ \hline
gDRH (128-bits)    & 0.425     &       0.531  \\ \hline
gDRH (256-bits)    & 0.583     &       0.629  \\ \hline
gDRH (512-bits)    & 0.668     &         0.724         \\ \hline
gDRH (1024-bits)   & 0.748    &       0.773         \\ \hline
gDRH (4096-bits)   & 0.783     &       0.815 \\ \hline
Max Conv Feature 512  &         0.797   &       0.824         \\ \hline
\end{tabular}
\end{table}

\subsubsection{The Effect of Region Proposals} 

In this sub-experiment, we explore firstly the two ways of region proposal: sliding window based and RPN based approaches. 
Both of them can generate different number of object proposals. For RPN, we choose $300$ regions to generate region hash code for lDRH. For sliding window, we tune $\lambda=[0.4,0.5,0.6,0.7]$ to get different number of object proposals. We further study the mAP of different top $M$ results.
gDRH can get some initial retrieval results, and in this subsection, we refine the results by using lDRH, i.e., gDRH+lDRH. The reported results (Tab.~\ref{RPN_SL}) are on the Oxford $5k$ dataset. From these results, we can make several observations. 

\begin{itemize}
\item Compared with RPN based DRH, the sliding window based DRH performs slightly better. It also requires less number of region boxes and does not need training a RPN network, thus we adopt sliding window based DRH as the default approach to conduct the following experiments. RPN is supposed to generate better region proposals than sliding window, but a worse performance is achieved in the experiment. One potential reason is that RPN is trained on pascal\_voc dataset, which is robust to propose the object of `buildings'. On the other hand, the objects in `Oxford buildings' are usually very large, thus they can be readily captured by sliding windows.

\item For sliding window based DRH, $\lambda$ affects the performance. In general, the performance is not very sensitive to $\lambda$. The best performance is achieved when $\lambda=0.6$, and it only generates about $40$ regions for each image. In the following experiments, the default setting for $\lambda$ is $0.6$.

\item  The top $M$ results have no significant impact on mAP either for DRH. In general, with the increase of $M$, the performance increases as well. When $M=800$, the algorithm achieves the best results. When $M=400$, the performance is $0.3\%$ lower than $M=800$. Considering the memory cost, we choose $M=400$ for the following experiments.
\end{itemize}

\begin{table}[]
\centering
\caption{The influence of Top $M$, $\lambda$ and the way of region generation. This experiment is conducted on the oxford $5k$ dataset with $L=1024$.}
\label{RPN_SL}
\resizebox{0.5\textwidth}{!}{
\begin{tabular}{|c|c|c|c|c|c|}
\hline
      & \multicolumn{4}{c|}{\textbf{Sliding Window based DRH}}     & \textbf{RPN-DRH} \\ \hline
\textbf{gDRH+lDRH}  & \begin{tabular}[c]{@{}c@{}}$\lambda=0.4$\\ $\sim$21boxes\end{tabular} & \begin{tabular}[c]{@{}c@{}}$\lambda=0.5$\\ $\sim$36 boxes\end{tabular} & \begin{tabular}[c]{@{}c@{}}$\lambda=0.6$\\ $\sim$40boxes\end{tabular} & \begin{tabular}[c]{@{}c@{}}$\lambda=0.7$\\ $\sim$60boxes\end{tabular} & $\sim$300 boxes     \\ \hline
\textbf{M=100} & 0.773   & 0.774    & 0.776   & 0.772   &   0.772      \\ \hline
\textbf{M=200} & 0.779   & 0.780    & 0.781   & 0.778   &     0.778    \\ \hline
\textbf{M=400} & 0.779   & 0.781    & \textbf{0.783}   & 0.780   &     0.780    \\ \hline
\textbf{M=800} & 0.782   & 0.784    & \textbf{0.786}  & 0.783   &     \textbf{0.783}    \\ \hline
\end{tabular}
}
\end{table}

\begin{table}[t]
	\centering
	\caption{The effect of query expansions (gQE and lQE).}
	\label{QE}
		\begin{tabular}{|@{}c@{}|@{}c|@{}c|@{}c|@{}c|@{}c|@{}c|}
			\hline
			\textbf{Settings}       & \textbf{gDRH} & \textbf{gQE} & \textbf{lDRH} & \textbf{lQE} & \textbf{\begin{tabular}[c]{@{}c@{}}oxford\\ 5k\end{tabular}} & \multicolumn{1}{c|}{\textbf{\begin{tabular}[c]{@{}c@{}}Paris\\ 6k\end{tabular}}} \\ \hline
			\multirow{2}{*}{\textbf{No QE}}          & yes  & no    & no   & no    & 0.745   &   0.773   \\ \cline{2-7} 
			& yes  & no  & yes  & no    & \textbf{0.783}    &  \textbf{0.801}     \\ \hline
			\multirow{4}{*}{\textbf{\begin{tabular}[c]{@{}c@{}} gQE \\ (different q)\end{tabular}}} &yes & yes,q=4        & no   & no    & 0.813  &   0.823   \\ \cline{2-7} 
			&yes& yes,q=5        & no   & no    &0.809         &   0.828   \\ \cline{2-7} 
			&yes& yes,q=6        & no   & no    &   \textbf{0.815}       &0.835      \\ \cline{2-7} 
			&yes& yes,q=7        & no   & no    &   0.789       & \textbf{0.842}     \\ \hline
			\multirow{4}{*}{\textbf{\begin{tabular}[c]{@{}c@{}}gQE+lQE\\ (q=6) \end{tabular}}} & yes& yes   & no   & no    & 0.815    &        0.835    \\ \cline{2-7} 
			&yes& yes   & yes  & no    & 0.804    & 0.831     \\ \cline{2-7} 
			&yes& no    & yes  & yes   & 0.833    &  0.819    \\ \cline{2-7} 
			&yes& yes   & yes  & yes   & \textbf{0.851}           &  0.849    \\ \hline

			\multirow{4}{*}{\textbf{\begin{tabular}[c]{@{}c@{}} gQE+lQE\\ (q=7) \end{tabular}}} &yes & yes   & no   & no    & 0.789    &        0.842    \\ \cline{2-7} 
			&yes& yes   & yes  & no    & 0.804    & 0.834     \\ \cline{2-7} 
			&yes& no    & yes  & yes   & 0.826    &  0.821    \\ \cline{2-7} 
			&yes& yes   & yes  & yes   & 0.838          &  \textbf{0.854}   \\ \hline
		\end{tabular}
\end{table}

\subsubsection{The Effect of Query Expansions} 
In this subsection, we study the effect of query expansions. This study aims to explore: the effect of lDRH, the effect of $q$ for gQE, the effect of gQE and lQE for DRH, as well as several combinations. The experiment results are shown in Tab.~\ref{QE}. From Tab.~\ref{QE}, we have some observations below:
\begin{itemize}
\item The lDRH improves the INS. In terms of mAP, for oxford $5k$ and paris $6k$ datasets, it increases by $3.8\%$ and $2.8\%$ , respectively.
\item The gQE improves the performance of DRH. When $q=6$, it performs the best (i.e., 85.1\%) on the Oxford $5k$ dataset, and when $q=7$ it gains the best performance on the Paris $6k$ dataset. In the following experiments, $q$ is set as $6$.
\item The experimental results show that by combining gQE and lQE, DRH performs the best (i.e., 85.1\% for oxford $5k$ and $85.4\%$ for Paris $6k$ datasets). Therefore, query expansion strategy can improve INS.
\end{itemize}

\begin{table}[t]
	\centering
	\caption{Comparison to state-of-the-art CNN representation.}
	\label{comparison}
	\resizebox{0.5\textwidth}{!}{
		\begin{tabular}{|l|l|c|c|c|c|}
			\hline
			& \multicolumn{1}{c|}{\textbf{Methods}}      & \textbf{\begin{tabular}[c]{@{}c@{}}Oxford \\ 5K\end{tabular}} & \textbf{\begin{tabular}[c]{@{}c@{}}Oxford\\ 105K\end{tabular}} & \textbf{\begin{tabular}[c]{@{}c@{}}Paris\\ 6K\end{tabular}} & \textbf{\begin{tabular}[c]{@{}c@{}}Paris\\ 106K\end{tabular}} \\ \hline
			\multirow{4}{*}{\begin{tabular}[c]{@{}l@{}}CNN \\ Feature\\ Encod-\\ing\end{tabular}} & Iscen et al.-Keans \cite{DBLP:journals/corr/IscenFGRJ14}       & 0.656     & 0.612      & 0.797   & 0.757     \\ \cline{2-6} 
			& Iscen et al.-Rand \cite{DBLP:journals/corr/IscenFGRJ14} & 0.251     & 0.437      & 0.212   & 0.444     \\ \cline{2-6} 
			& Ng et al. \cite{DBLP:journals/corr/NgYD15}         & 0.649     & - & 0.694   & -         \\ \cline{2-6} 
			& Iscen et al. \cite{Iscen_2016_CVPR}       & 0.737     & 0.655      & 0.853   & 0.789     \\ \hline
			\multirow{8}{*}{\begin{tabular}[c]{@{}l@{}}CNN\\ Featues\\ Aggre-\\gation\end{tabular}}        & Razavian et al. \cite{DBLP:journals/corr/RazavianSMC14}    & 0.556     & - & 0.697   & -         \\ \cline{2-6} 
			& Bebenko et al. \cite{DBLP:journals/corr/BabenkoL15}      & 0.657     & 0.642      & -       & -         \\ \cline{2-6} 
			& Tolias et al. \cite{DBLP:journals/corr/ToliasSJ15}      & 0.669     & 0.616      & 0.830   & 0.757     \\ \cline{2-6} 
			& Kalantidis et al. \cite{DBLP:journals/corr/KalantidisMO15} & 0.684     & 0.637      & 0.765   & 0.691     \\ \cline{2-6} 
			& Mohedano et al. \cite{DBLP:journals/corr/MohedanoSMMON16}    & 0.739     & 0.593      & 0.820   & 0.648     \\ \cline{2-6} 
			& Salvador et al. \cite{Salvador_2016_CVPR_Workshops}   & 0.710     & - & 0.798   & -         \\ \cline{2-6} 
			& Tao et al. \cite{cvpr20146909666}        & 0.765     & - & -       & -         \\ \cline{2-6} 
			& Tao et al.  \cite{7298613}        & 0.722     & - & -       & -         \\ \hline
			\multirow{4}{*}{\begin{tabular}[c]{@{}l@{}}Integra\\-tion\\ with \\ Locaility \\ and \\ QE\end{tabular}} & \begin{tabular}[c]{@{}l@{}}Kalantidis et al.\\ +GQE \cite{DBLP:journals/corr/KalantidisMO15} \end{tabular}      & 0.749     & 0.706      & 0.848   & 0.794     \\ \cline{2-6} 
			& \begin{tabular}[c]{@{}l@{}}Tolias et al.\\ +AML+QE \cite{DBLP:journals/corr/ToliasSJ15}\end{tabular}       & 0.773     & 0.732      & \multicolumn{1}{l|}{\textbf{0.865}}       & \multicolumn{1}{l|}{0.798}         \\ \cline{2-6} 
			& \begin{tabular}[c]{@{}l@{}}Mohedano et al. \cite{DBLP:journals/corr/MohedanoSMMON16}\\ +Rerank+LQE\end{tabular} & 0.788     & 0.651      & 0.848   & 0.641     \\ \cline{2-6} 
			& \begin{tabular}[c]{@{}l@{}}Salvador et al. \cite{Salvador_2016_CVPR_Workshops}\\ +CS-SR+QE\end{tabular}   & 0.786     & - & 0.842   & -         \\ \hline
			\multirow{3}{*}{Hashing}     & Jegou et al. \cite{JegouDS08}    & \multicolumn{1}{c|}{0.503}         & \multicolumn{1}{l|}{-}          & \multicolumn{1}{c|}{0.491}       & \multicolumn{1}{c|}{-}         \\ \cline{2-6} 
			& Liu et al. \cite{LiuLZZT14}     & \multicolumn{1}{c|}{0.518}         & \multicolumn{1}{l|}{-}          & \multicolumn{1}{l|}{0.511}       & \multicolumn{1}{c|}{-}         \\ \hline        
			\multirow{3}{*}{ours}     & DRH(gDRH+lDRH)    & \multicolumn{1}{c|}{0.783}         & \multicolumn{1}{c|}{0.754}          & \multicolumn{1}{l|}{0.801}       & \multicolumn{1}{l|}{0.733}         \\ \cline{2-6} 
			& DRH All     & \multicolumn{1}{c|}{\textbf{0.851}}         & \multicolumn{1}{c|}{\textbf{0.825}}          & \multicolumn{1}{c|}{0.849}       & \multicolumn{1}{c|}{\textbf{0.802}}         \\ \cline{2-6} 
			\eat{
				& DRH All(q=7)      & \multicolumn{1}{c|}{0.838}         & \multicolumn{1}{l|}{0.822}      & \multicolumn{1}{l|}{0.854}       & \multicolumn{1}{l|}{0.809}     \\ 
			}\hline
		\end{tabular}
	}
\end{table}
\subsection{Comparing with State-of-the-art Methods}
To evaluate the performance of DRH, we compare our methods with the state-of-the-art deep features based methods. These methods can be divided into four categories:

\begin{itemize}
\item Feature Encoding approaches \cite{DBLP:journals/corr/IscenFGRJ14,Iscen_2016_CVPR,DBLP:journals/corr/NgYD15}. They aim to reducing the number of vectors against which the query is compared. \cite{Iscen_2016_CVPR} designed a dictionary learning approach for global descriptors obtained from both SIFT and CNN features. \cite{DBLP:journals/corr/NgYD15} extracted convolutional features from different layers of the networks, and adopted VLAD encoding to encode features into a single vector for each image. 

\item Aggregating Deep features \cite{DBLP:journals/corr/RazavianSMC14,DBLP:journals/corr/BabenkoL15,DBLP:journals/corr/ToliasSJ15,DBLP:journals/corr/KalantidisMO15,DBLP:journals/corr/MohedanoSMMON16,Salvador_2016_CVPR_Workshops,cvpr20146909666,7298613}. \cite{DBLP:journals/corr/RazavianSMC14,DBLP:journals/corr/BabenkoL15,DBLP:journals/corr/ToliasSJ15,DBLP:journals/corr/KalantidisMO15} are focusing on aggregating deep convolutional features for image or instance retrieval. In \cite{DBLP:journals/corr/MohedanoSMMON16}, Mohedano \textit{et al.} propose a simple instance retrieval pipeline based on encoding the convolutional features of CNN using the bag of words aggregation scheme (BoW). \cite{Salvador_2016_CVPR_Workshops} explores the suitability for instance retrieval of image- and region-wise descriptors pooled from Faster RCNN. For \cite{cvpr20146909666,7298613}, they are focusing on generic instance search. 

\item Integrating with Locality and QE. Some of the previous methods are extended by integrating re-ranking and/or query expansion techniques \cite{DBLP:journals/corr/KalantidisMO15,DBLP:journals/corr/ToliasSJ15,DBLP:journals/corr/MohedanoSMMON16,Salvador_2016_CVPR_Workshops}. 

\item Hashing methods. Lots of hashing methods are proposed for image retrieval \cite{JegouDS08,LiuLZZT14}.
\end{itemize}


The comparison results are shown in Tab.~\ref{comparison} and we also show some qualitative results in Fig.~\ref{fig:example}. From these results, we have the following observations. 
\begin{itemize}
	\item First, our approach performs the best on both two large-scale datasets oxford $105k$ and Paris $106k$ with $0.825$ and $0.802$, respectively. In addition, the performance on oxford $105k$ is higher than Tolias \textit{et al.} +AML+QE \cite{DBLP:journals/corr/ToliasSJ15} with a $9.3\%$ increase. 
	\item On Paris $6k$ dataset, our performance is slightly lower than that of Tolias \textit{et al.} +AML+QE \cite{DBLP:journals/corr/ToliasSJ15} by $1.1\%$, but on the Oxford $5k$ dataset, it outperforms Mohedano \textit{et al.}+Rerank+LQE \cite{DBLP:journals/corr/MohedanoSMMON16} by $6.3\%$.
	\item The performance of CNN features aggregation is better than CNN features in general, and it can be further improved by integrating re-ranking and query expansion strategies. On the other hand, hashing methods performs the worst in terms of mAP. This is probably due to the information loss of hash codes.
	\item The qualitative results show that DRH can retrieve instances precisely, even for the cases when the query image is different from the whole target image, but similar to a small region of it.
\end{itemize}

\subsection{Efficiency Study}
In this subsection, we compare the efficiency of our DRH to the state-of-the-art algorithms.
The time cost for instance search usually consists of two parts: filtering and re-ranking. For our DRH, the re-ranking is conducted using M=400 candidates, each has around 40 local regions. Therefore, the time cost for re-ranking can be ignored compared with the filtering step. For the comparing algorithms, they have different re-ranking strategies, and some of them do not have this step. Therefore, we only report the time for filtering step of these comparing algorithms.

We choose \cite{DBLP:journals/corr/ToliasSJ15} as the representative algorithm for non-hashing algorithms. To make fair comparison, we implement the linear scan of the conv5 feature \cite{DBLP:journals/corr/ToliasSJ15} and our hash codes using Python and C++ respectively, without any accelerating strategies. All the experiments are done on a server with Intel@Core i7-6700K CPU@ 4.00GB$\times$8. The graphics is GeGorce GTX TITAN X/PCle/SSE2. The search time is average number of milliseconds to return top $M=400$ results to a query, and the time costs are reported in Tab. \ref{tab.time}.

\begin{table}[t]
	\centering
	\small
	\caption{The time cost (ms) for different algorithms}
	\label{tab.time}
	\begin{tabular}{|c|c|c|c|}\hline
		Datasets     & \begin{tabular}[c]{@{}c@{}}CNN feature\cite{DBLP:journals/corr/ToliasSJ15}\\
			 512-D conv5\end{tabular} & \begin{tabular}[c]{@{}c@{}}DRH\\ 512-D hash\end{tabular} & \begin{tabular}[c]{@{}c@{}}DRH\\ 1024-D hash\end{tabular} \\ \hline\hline
		Oxford 105K & 1078    & 3     & 12     \\\hline
		Paris 106K  & 1137    & 3  & 12               \\\hline                     
	\end{tabular}
\end{table}

There results show that directly using max pooling features extracted from conv5 to conduct INS search is much slower than using our hash codes. It takes $1078$ ms on Oxford $105k$ dataset and $1137$ ms on Paris $106k$ dataset for \cite{DBLP:journals/corr/ToliasSJ15}. By contrast, when the code length is $512$, our DRH is more than 300 times faster than \cite{DBLP:journals/corr/ToliasSJ15}, and it takes $3$ ms on both Oxford $105k$ and Paris $106k$ datasets. Even when the code length increases to 1024, DRH can achieve nearly 100 times faster than \cite{DBLP:journals/corr/ToliasSJ15}. Therefore, our method has the ability to work on large scale datasets.

\section{Conclusion}
In this work, we propose an unsupervised deep region hashing (DRH) method, a fast and efficient approach for large-scale INS with an image patch as the query input. We firstly utilize deep conv feature map to extract nearly $40$ to generate hash codes to represent each image. Next, the gDRH search, gQE, lDRH and lQE are applied to further improve the INS search performance. Experiment on four datasets demonstrate the superiority of our DRH compared to others in terms of both efficiency and effectiveness.

\end{document}